\documentclass[journal]{IEEEtran}

\usepackage[table]{xcolor}

\usepackage{cite}
\usepackage{amsmath,amssymb,amsfonts}
\usepackage{algorithmic}
\usepackage{graphicx}
\usepackage{textcomp}

\begin{document}

\title{ESearch-R1: Learning Cost-Aware MLLM Agents for Interactive Embodied Search via Reinforcement Learning}

\author{Weijie~Zhou,
        Xuangtang~Xiong,
        Ye~Tian,
        Lijun~Yue,
        Xinyu~Wu,
        Wei~Li,
        Chaoyang~Zhao,
        Honghui~Dong*,
        Ming~Tang,
        Jinqiao~Wang,
        and~Zhengyou~Zhang,
\thanks{*Corresponding author: Honghui Dong.}%
\thanks{Weijie Zhou, Wei Li, Honghui Dong, and Jinqiao Wang are with the School of Traffic and Transportation, Beijing Jiaotong University, Beijing, China.}%
\thanks{Xuangtang Xiong, Ye Tian, Lijun Yue, and Zhengyou Zhang are with Tencent Robotics X \& Futian Laboratory, Shenzhen, China.}%
\thanks{Chaoyang Zhao, Ming Tang, and Jinqiao Wang are with the Foundation Model Research Center, Institute of Automation, Chinese Academy of Sciences, Beijing, China.}%
\thanks{Xinyu Wu is with the University of Chinese Academy of Sciences, Beijing, China.}%
}

\maketitle

\begin{abstract}
Multimodal Large Language Models (MLLMs) have empowered embodied agents with remarkable capabilities in planning and reasoning. However, when facing ambiguous natural language instructions (e.g., "fetch the tool" in a cluttered room), current agents often fail to balance the high cost of physical exploration against the cognitive cost of human interaction. They typically treat disambiguation as a passive perception problem, lacking the strategic reasoning to minimize total task execution costs.
To bridge this gap, we propose ESearch-R1, a cost-aware embodied reasoning framework that unifies interactive dialogue (Ask), episodic memory retrieval (GetMemory), and physical navigation (Navigate) into a single decision process. We introduce HC-GRPO (Heterogeneous Cost-Aware Group Relative Policy Optimization). Unlike traditional PPO which relies on a separate value critic, HC-GRPO optimizes the MLLM by sampling groups of reasoning trajectories and reinforcing those that achieve the optimal trade-off between information gain and heterogeneous costs (e.g., navigate time, and human attention). 
Extensive experiments in AI2-THOR demonstrate that ESearch-R1 significantly outperforms standard ReAct-based agents. It improves task success rates while reducing total operational costs by approximately 50\%, validating the effectiveness of GRPO in aligning MLLM agents with physical world constraints.

\end{abstract}


\section{Introduction}
\label{sec:introduction}

\begin{figure}[t]
\centering
\includegraphics[width=\linewidth]{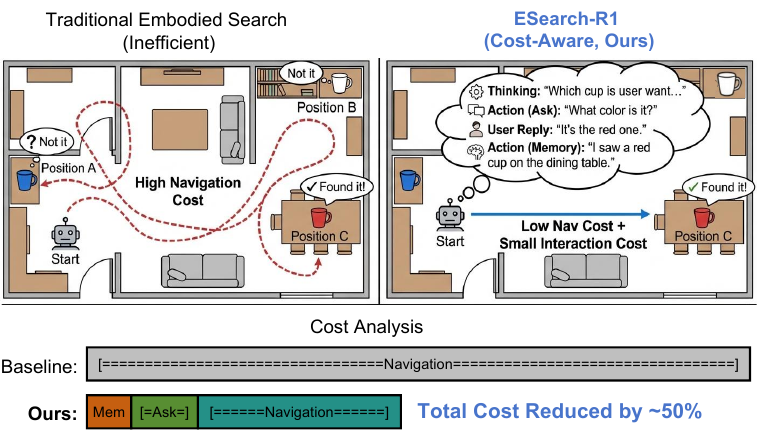}
\caption{From passive search to proactive interactive disambiguation. In scenes with multiple candidate instances, traditional agents rely on exhaustive navigation and inspection, incurring high movement cost. ESearch-R1 behaves as a collaborative partner: it first narrows down the candidate set via asking the user and retrieving episodic memory, and then executes the shortest path to the uniquely identified goal, thereby reducing both navigation cost and the number of user interruptions.}
\label{fig:paradigm}
\end{figure}

The integration of Multimodal Large Language Models (MLLMs) into robotic systems has catalyzed a shift from rigid instruction following to general-purpose embodied agents capable of long-horizon planning \cite{planning3, planning4, embodiedrail, lightplanner, physvlm, physvlmavr}. While these agents demonstrate impressive semantic understanding, a critical misalignment remains between the MLLM's digital training objective (next-token prediction) and the physical reality of robotic operation \cite{active_object_search, agent1, agent2, embodiedagent1, embodiedagent2, embodiedagent3}.

In the digital world, generating a thought is cheap; in the physical world, execution is expensive. A major challenge arises when agents face ambiguity—such as under-specified instructions (e.g., "Find the cup" in a room with multiple cups) \cite{ambiguous1, ambiguous2, ask-to-act}. Standard MLLM agents \cite{perceptionr1, webagent, ominir1}, often relying on ReAct \cite{react} or zero-shot prompting, tend to hallucinate a specific target or resort to "brute-force" physical search \cite{embodied-search-habitat, embodied-resoner}, sequentially inspecting every candidate. This passive exploration strategy incurs prohibitive costs in terms of time, energy, and wear-and-tear. Conversely, overly cautious agents might spam the user with trivial questions, degrading the human interaction experience.

We argue that resolving ambiguity is not merely a perception task but a cost-aware reasoning challenge. A truly intelligent embodied agent should possess "System 2" thinking capabilities: explicitly weighing the high cost of physical movement against the low cost of internal memory retrieval or the moderate cost of asking a clarification question. The agent must learn a meta-policy that trades off information gain against these heterogeneous costs before taking a step.

To achieve this, we introduce ESearch-R1, a novel framework that aligns MLLM reasoning with physical constraints. We formulate the interactive search problem as a Partially Observable Markov Decision Process (POMDP) where dialogue, memory, and navigation are unified actions with distinct cost profiles.

The core contribution of ESearch-R1 is its learning paradigm. We propose HC-GRPO (Heterogeneous Cost-Aware Group Relative Policy Optimization). Instead of training a complex value network (Critic) as in standard PPO—which is often unstable for Large Language Models—HC-GRPO leverages group-based sampling. For each query, the agent generates a group of reasoning trajectories (Chain-of-Thought) and actions. The optimization step then reinforces trajectories that maximize a trajectory-level hybrid reward, specifically penalizing high-cost physical actions when cheaper information-seeking alternatives (like referencing episodic memory) were available.

This approach enables the emergence of sophisticated disambiguation strategies without explicit rule-coding. The agent learns to "think before it moves": checking its internal memory first, asking a targeted question if necessary, and navigating only when the target is uniquely identified. We validate ESearch-R1 on ESearch-Bench, a benchmark built on AI2-THOR representing unstructured semantic environments. Experiments show that our GRPO-tuned agent outperforms strong ReAct baselines (including larger models like Gemini-Pro) by reducing total task costs by $\sim$50\%, effectively bridging the gap between digital reasoning and physical efficiency.
Our main contributions are:
\begin{itemize}
\item \textbf{A Cost-Aware Embodied Decision Framework (ESearch-R1)} that unifies dialogue, episodic memory, and navigation, enabling agents to actively resolve ambiguity.
\item \textbf{Heterogeneous Cost-Aware GRPO (HC-GRPO)}, a reinforcement learning algorithm tailored for MLLMs that eliminates the need for a critic network. It effectively optimizes reasoning chains (CoT) to minimize the joint cost of navigation and human interaction.
\item \textbf{Emergent Efficient Behaviors.} We empirically demonstrate that cost-sensitive RL induces the agent to substitute expensive physical exploration with low-cost cognitive retrieval and strategic dialogue, achieving state-of-the-art efficiency in simulation.
\end{itemize}

\section{Related Work}

Our work sits at the intersection of embodied visual search, active perception, and the alignment of MLLM agents via reinforcement learning. We review these areas to highlight the shift from passive instruction following to cost-aware embodied reasoning, positioning ESearch-R1 within the recent advancements in reasoning models.

\subsection{Visual Search and Navigation under Uncertainty}

Visual search is a cornerstone of embodied AI \cite{mmsearchr1, mmsearch2, mmsearch, embodied-resoner, ask-to-act}. Standard benchmarks such as Object Navigation (ObjectNav) \cite{obejctnav, objectnav2, objectnav3} and Instance Navigation (InstanceNav) \cite{instancenav, instancenav2} evaluate an agent's ability to locate objects based on semantic categories or specific queries. While recent methods leverage hierarchical mapping and frontier-based exploration to improve efficiency, they predominantly model the problem as passive execution.

A critical gap in current literature is the handling of ambiguity in natural language instructions (e.g., "fetch the tool" in a garage with multiple tools) \cite{ambiguous1, ask-to-act, ambiguous2}. Most existing baselines operate under a "trust-and-execute" paradigm, assuming the target is uniquely identifiable or relying on exhaustive, brute-force search when it is not. From a reasoning perspective, this represents a failure of "System 2" thinking: the agent physically explores the environment to resolve uncertainty that could often be resolved more cheaply via cognitive operations (e.g., retrieving episodic memory). ESearch-R1 addresses this by treating navigation not just as a path-planning problem, but as an information-seeking process where physical movement is a high-cost action to be minimized.

\subsection{Interactive Agents and Active Perception}

To overcome the limitations of passive search, recent works have explored interactive agents capable of querying humans for assistance \cite{teach, ask-to-act, asking2}. Benchmarks like TEACh \cite{teach} and DialFRED \cite{dialfred} introduced scenarios where agents must engage in dialogue to complete tasks. Approaches such as Ask-To-Act \cite{ask-to-act} use reinforcement learning or uncertainty estimation to decide when to request help.

However, prior works typically treat dialogue as an isolated module or a discrete fallback mechanism, lacking a unified heterogeneous cost model. They often fail to weigh the "social cost" of disturbing a human against the "physical cost" of navigation. Moreover, they rarely integrate episodic memory retrieval as a competing alternative to dialogue and exploration. In contrast, ESearch-R1 formulates these diverse capabilities—Ask, GetMemory, and Navigate—within a single reasoning action space. By doing so, our agent learns to actively prune the search space using the cheapest available information source, mimicking human-like cost-effective disambiguation strategies.

\subsection{Aligning MLLM Agents via Reinforcement Learning}

The advent of Multimodal Large Language Models (MLLMs) has empowered agents with strong semantic reasoning and zero-shot planning capabilities \cite{planning3, planning4, embodiedrail}. Frameworks like ReAct \cite{react} enable agents to interleave thought and action. Despite their success, standard MLLMs suffer from a misalignment between their pre-training objective (next-token prediction) and the physical constraints of embodied tasks. They often exhibit "cost-blindness," treating a computationally cheap text generation step and an energetically expensive robot movement as equivalent.

While Reinforcement Learning (RL) allows for fine-tuning models on task rewards, applying standard algorithms like PPO to multi-billion parameter models introduces significant instability, primarily due to the difficulty of training an accurate value critic for complex reasoning traces \cite{rl3, rl4, rl5}. Inspired by recent successes in reasoning models \cite{deepseekr1, rl1, rl2, team2025kimi, kimi2}, our work leverages Group Relative Policy Optimization (GRPO). Unlike traditional PPO which relies on a separate value network, GRPO optimizes the policy by contrasting a group of sampled reasoning trajectories against each other.

In ESearch-R1, we adapt this paradigm to HC-GRPO (Heterogeneous Cost-Aware GRPO). This approach allows the MLLM to self-reinforce efficient behaviors—learning to "think" (retrieve memory) before "acting" (navigating)—effectively grounding the abstract reasoning of LLMs into the cost-constrained reality of the physical world.

\begin{figure*}[t]
    \centering
    \includegraphics[width=\linewidth]{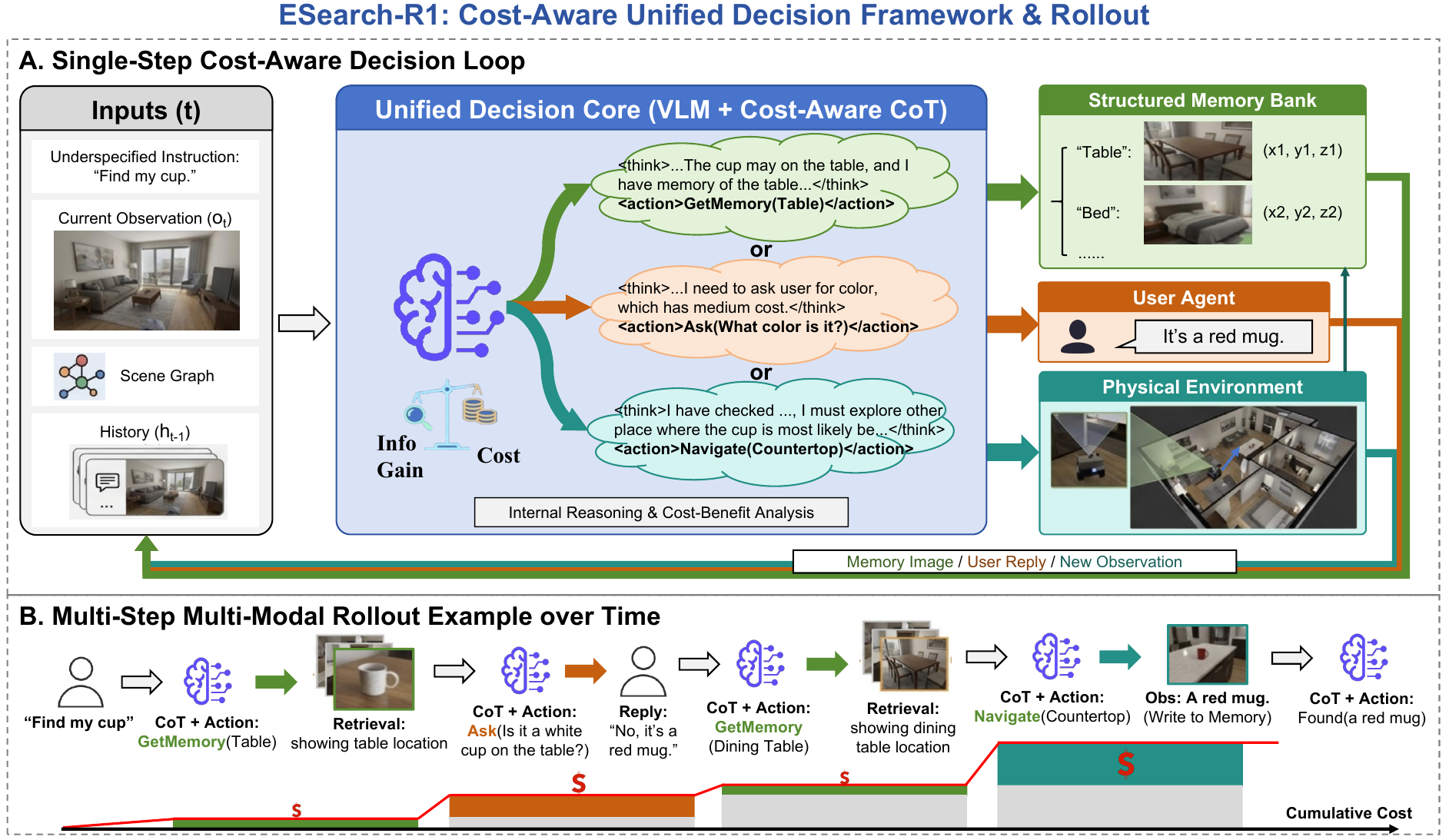}
    \caption{ESearch-R1 architecture and closed-loop inference. Under multimodal context, the unified decision core assigns comparable information–cost trade-offs to Ask/GetMemory/Navigate; new observations from exploration are written into memory, forming a closed perception–memory–decision loop.}
    \label{fig:framework}
\end{figure*}

\section{ESearch-R1: Cost-Aware Embodied Reasoning Framework}
\label{sec:method}

This section details the proposed ESearch-R1 framework. We first present the cognitive architecture built upon a Multimodal Large Language Model (MLLM). Then, we formalize the problem as a Cost-Aware Partially Observable Markov Decision Process (POMDP). Finally, we introduce HC-GRPO, a group-relative reinforcement learning algorithm designed to align reasoning capabilities with physical cost constraints.

\subsection{Cognitive Agent Architecture}
The core of ESearch-R1 is a cognitive agent that integrates multimodal perception, semantic memory, and a unified reasoning interface. The agent operates in a closed perception-action loop to progressively reduce target uncertainty.

\subsubsection{Unified Reasoning Core}
We leverage a pre-trained Multimodal Large Language Model (MLLM) as the central policy $\pi_\theta$. At each time step $t$, the MLLM processes a multimodal context $h_t$, outputs a \textit{Chain-of-Thought (CoT)} reasoning trace, and executes an action $a_t$. This enables the agent to deliberate—weighing information value against acquisition cost—before acting.

\subsubsection{Unified Action Space}
We define a hybrid action space $\mathcal{A}$ that unifies communicative, cognitive, and physical capabilities:
\begin{itemize}
    \item \texttt{Navigate(loc)}: A high-cost physical action moving the agent to a semantic location to acquire visual info.
    \item \texttt{Ask(query)}: A communicative action to query the user, incurring a social cost (cognitive load).
    \item \texttt{GetMemory(key)}: A cognitive action retrieving past experiences, incurring negligible cost.
    \item \texttt{Found(target)}: A terminal action to conclude the task.
\end{itemize}

\subsection{Problem Formulation: Cost-Aware POMDP}
We formulate the interactive search as a POMDP defined by $\langle \mathcal{S}, \mathcal{A}, \mathcal{O}, \mathcal{R}, \mathcal{C}, \gamma \rangle$.

\subsubsection{Heterogeneous Cost Function}
A critical component is the cost function $C(a_t, h_t)$, explicitly modeling the disparate nature of resources. We define it as:
\begin{equation}
    C(a_t) = 
    \begin{cases} 
        c_{\text{nav}} \cdot d(p_t, p_{t+1}) & \text{if } a_t = \texttt{Nav} \\
        c_{\text{ask}} \cdot (1 + \alpha N_{\text{ask}}) & \text{if } a_t = \texttt{Ask} \\
        c_{\text{mem}} & \text{if } a_t = \texttt{Mem}
    \end{cases}
    \label{eq:cost}
\end{equation}
where $d(\cdot)$ denotes physical distance, $N_{\text{ask}}$ is the cumulative number of questions, and $\alpha$ is a fatigue penalty factor. We enforce $c_{\text{nav}} > c_{\text{ask}} \gg c_{\text{mem}}$, reflecting that internal recall is cheap while physical movement is expensive.

\subsubsection{Objective}
The agent aims to maximize the expected return $J(\theta) = \mathbb{E}_{\tau \sim \pi_\theta} [R(\tau)]$, where the trajectory reward is:
\begin{equation}
    R(\tau) = R_{\text{task}} - \lambda \sum_{t=0}^{T} C(a_t)
\end{equation}
Here, $R_{\text{task}}$ is a sparse success reward, and $\lambda$ balances success against efficiency.

\subsection{HC-GRPO: Heterogeneous Cost-Aware Optimization}
Standard PPO is often unstable for reasoning-heavy MLLMs due to the difficulty of training a reliable value critic. We propose \textbf{HC-GRPO} (Heterogeneous Cost-Aware Group Relative Policy Optimization), which eliminates the critic by leveraging group-based advantage estimation.

\subsubsection{Group Relative Advantage}
For each instruction $q$, the policy generates a group of $G$ outputs $\{o_1, \dots, o_G\}$. We compute the advantage $A_i$ of each output $o_i$ relative to the group statistics:
\begin{equation}
    A_i = \frac{r_i - \mu_{R}}{\sigma_{R} + \epsilon}
    \label{eq:advantage}
\end{equation}
where $r_i$ is the reward of trajectory $i$, and $\mu_{R}, \sigma_{R}$ are the mean and standard deviation of rewards within the group. This encourages trajectories that achieve the task with lower costs than their peers.

\subsubsection{Optimization Objective}
The policy is optimized to maximize the surrogate objective subject to a KL-divergence constraint. To fit the double-column layout, we formulate the loss $\mathcal{L}(\theta)$ as:
\begin{equation}
\begin{split}
    \mathcal{L}(\theta) = \mathbb{E}_{q \sim D} \Bigg[ \frac{1}{G} \sum_{i=1}^{G} \bigg( & \min \left( \rho_i A_i, \text{clip}(\rho_i, 1-\epsilon, 1+\epsilon) A_i \right) \\
    & - \beta D_{KL}(\pi_\theta || \pi_{\text{ref}}) \bigg) \Bigg]
\end{split}
\label{eq:grpo}
\end{equation}
where $\rho_i = \frac{\pi_\theta(o_i|q)}{\pi_{\text{old}}(o_i|q)}$ is the importance sampling ratio, and $\pi_{\text{ref}}$ is the SFT warm-up policy. 

\subsubsection{Emergent Reasoning}
By optimizing Eq. \ref{eq:grpo} with the cost-aware reward (Eq. \ref{eq:cost}), the agent learns to associate "thinking" (Cost $\approx 0$) with higher relative rewards compared to "acting blindly" (Cost $\gg 0$). This induces emergent behaviors where the agent checks memory first, effectively functioning as a "System 2" planner.

\section{ESearch-Bench: A Benchmark for Embodied Reasoning under Uncertainty}
\label{sec:benchmark}

Evaluating an agent's ability to trade off physical exploration against cognitive interaction requires a specialized testing ground. Most existing benchmarks (e.g., ObjectNav \cite{obejctnav}) assume precise goals or focus purely on navigation mechanics, failing to capture the semantic ambiguity inherent in real-world human instructions.

To address this, we introduce \textbf{ESearch-Bench}.
Unlike static datasets, ESearch-Bench serves as a dynamic evaluation platform where the core challenge is Cost-Aware Disambiguation. It features \textbf{under-specified instructions} (e.g., "Find the cup") in cluttered environments, forcing the agent to proactively reason about how to acquire missing information efficiently.

This section details our automated environment setup, the LLM-driven task generation pipeline, and the benchmark statistics.

\subsection{Simulation Environment: A Proxy for Semantic Complexity}
We build ESearch-Bench upon the AI2-THOR \cite{ai2thor} simulator. While typically used for household robotics, its high-fidelity physics and dense object placement make it an ideal proxy for general semantically complex environments (e.g., cluttered workshops, offices, or labs).

We utilize a diverse set of 120 interactive scenes, strictly partitioned into training (80 scenes) and testing (30 \textbf{unseen} scenes). This separation is crucial to evaluate the agent's ability to generalize its reasoning meta-policy—learning when to ask or look—rather than memorizing specific floor plans.

To isolate high-level reasoning from low-level control noise, we abstract the actuation into semantic primitives (e.g., \texttt{Navigate(object)}), implemented via a reliable path planner (A*) on the simulator's navigation mesh.

\subsection{Scalable Task Synthesis Pipeline}
Manually annotating ambiguous tasks is prohibitively expensive. We developed a fully automated pipeline leveraging a commercial MLLM (Gemini-1.5-Pro) to synthesize grounded, ambiguous instructions at scale. The process consists of three phases:

\subsubsection{Ambiguity Injection via Scene Graph Analysis}
We first parse the metadata of each scene to construct a Semantic Scene Graph. The system algorithmically identifies "Ambiguity Sets"—clusters of objects sharing the same coarse category (e.g., multiple \textit{mugs} or \textit{bottles} in a kitchen).

To simulate natural language under-specification, the MLLM is prompted to generate an instruction $I$ that applies to $N$ candidates ($2 \le N \le 5$) within the set. Crucially, only one specific instance is designated as the Ground Truth (GT) target, creating a "Needle in a Haystack" scenario that cannot be solved by random guessing.

\subsubsection{Enforcing Partial Observability}
Aligning with our POMDP formulation, we enforce strict partial observability. The agent initializes with only the ambiguous instruction $I$ and its egocentric view. The Scene Graph and target ID are completely redacted. This information asymmetry is the engine of the benchmark: it compels the agent to generate a Chain-of-Thought to determine whether to resolve the uncertainty via dialogue (Ask) or physical verification (Navigate).

\subsubsection{Simulated Human Oracle with Fatigue Modeling}
To enable closed-loop evaluation without human-in-the-loop latency, we implement a Simulated User (Oracle). This module answers the agent's queries based on the GT target's unique attributes (e.g., "It's the red one near the sink").

However, to model the real-world cost of human interruption, we introduce a Dynamic Cooperation Mechanism:

\begin{itemize}
\item \textbf{Decaying Helpfulness:} The oracle models "user fatigue." For the first query, the response is highly detailed. As the number of questions ($N_{ask}$) increases, the probability of receiving a useful answer decays exponentially (from 100
\item \textbf{Implicit Cost Signal:} This behavior acts as an implicit negative reward, teaching the GRPO-optimized agent that "asking is a limited resource," encouraging it to prioritize internal memory retrieval or targeted questions over spamming.
\end{itemize}

\subsection{Data Statistics and Expert Reasoning Traces}
\subsubsection{SFT Training Data with CoT}
To "warm start" our agent's reasoning capabilities, we generated 800 expert trajectories. Unlike standard behavioral cloning datasets that only contain (State, Action) pairs, we augment our data with Reasoning Traces.

An Oracle Planner (with access to global ground truth) solves each task by calculating the minimum-cost path. We then use an LLM to annotate the decision process with natural language rationales (e.g., \textit{"Reasoning: There are 4 candidates. Navigating to all of them is too far (Cost > 50). I should check my memory first..."}). These traces are critical for the SFT stage of our HC-GRPO framework.

\subsubsection{Evaluation Protocols}
The \textbf{ESearch-Bench} test set comprises 330 unique tasks in unseen environments. We categorize tasks by difficulty based on the size of the Ambiguity Set (Easy: 2 candidates; Hard: 4+ candidates).
Furthermore, to test zero-shot generalization, 15\% of test instructions involve object categories not seen during training. This rigorously evaluates whether the agent has learned a generalizable "search strategy" rather than object-specific priors.

\section{Experiments and Analysis}
\label{sec:experiments}

This section presents a comprehensive empirical evaluation of the proposed ESearch-R1 framework. Our experiments are designed to answer two key research questions relevant to embodied AI and human-robot collaboration:
\begin{itemize}
    \item RQ1 (Performance): Can the proposed framework effectively resolve instruction ambiguity in unstructured environments compared to existing methods?
    \item RQ2 (Cost Efficiency): Does the cost-aware mechanism successfully minimize the joint cost of human attention (dialogue) and robotic operation (navigation)?
\end{itemize}

\subsection{Implementation Details}

\paragraph{Environment and Models}
Our agent's cognitive core is based on the \texttt{Qwen2.5-VL-7B} multimodal large language model \cite{qwen2_5_vl}, which we fine-tune in all experiments. All experiments were conducted on servers equipped with 8$\times$NVIDIA H20 GPUs (96~GB each).

\paragraph{Stage I: Supervised Fine-Tuning (SFT-CoT)}
We performed supervised fine-tuning for one epoch using the AdamW optimizer. Key hyperparameters included a learning rate of $1\times10^{-5}$ with a cosine decay schedule (ratio $=0.1$) and a batch size of 16.

\paragraph{Stage II: Online Optimization via HC-GRPO}
The policy was trained online for three epochs. We employed a GRPO-style update rule with the following hyperparameters: learning rate $=2\times10^{-6}$, batch size $=8$, discount factor $\gamma=0.99$, KL divergence coefficient $\beta=0.1$, value loss weight $c_1=1.0$, entropy coefficient $c_2=0.01$, and GRPO clipping parameter $\epsilon=0.2$. The reference policy $\pi_{\text{ref}}$ was initialized from the SFT checkpoint and remained frozen. To handle malformed action generation, we assigned a small negative reward and allowed the model to retry once.

\paragraph{Cost and Reward Hyperparameters}
We set the terminal task rewards to $R_{\text{success}}=1.0$ and $R_{\text{fail}}=-0.1$, with a cost-sensitivity coefficient of $\lambda=1.0$. Per-action base costs were defined as: $c_{\text{nav}}=1.0$ (navigation), $c_{\text{ask\_base}}=0.5$ (dialogue), and $c_{\text{mem}}=0.01$ (memory retrieval). The dialogue cost increased with each query (growth factor $\alpha=0.2$), and the navigation cost was scaled by distance ($c_{\text{nav}} \cdot d_t$). A format penalty of $c_{\text{format}}=0.1$ was also applied.

\paragraph{Statistical Significance}
To ensure the robustness of our findings, all reported metrics are the mean $\pm$ standard deviation computed over five independent runs with different random seeds.

\subsection{Experimental Setup}

\paragraph{Simulation Environment as a Proxy}
We utilize the AI2-THOR simulator to construct ESearch-Bench. While visually represented as indoor scenes, this environment serves as a rigorous proxy for complex unstructured environments. It captures the core challenges of collaborative tasks: partial observability, semantic ambiguity, and the high cost of physical movement. The benchmark includes 330 tasks across 30 unseen environments to test generalization.

\paragraph{Baselines}
We compare against state-of-the-art interactive agents. To ensure a fair comparison of decision-making efficiency, all baselines operate within the same action space but employ different strategies:
\begin{itemize}
    \item \textbf{ReAct \cite{react} (Zero-shot) :} We instantiate the ReAct paradigm with several MLLMs: \texttt{Gemini-2.5-Flash} \cite{gemini2.5}, \texttt{Gemini-2.5-Pro} \cite{gemini2.5}, \texttt{Qwen2.5-VL-7B} \cite{qwen2_5_vl}, and \texttt{Qwen2.5-VL-32B} \cite{qwen2_5_vl}. The agent generates decisions via zero-shot "think-act" prompting.
    \item \textbf{Ask-To-Act \cite{ask-to-act} (Fine-tuned):} A strong baseline for interactive agents that can ask clarification questions. However, it does not explicitly model or optimize for the heterogeneous costs of dialogue versus physical actions, lacking a unified policy for cost-aware trade-offs. We adapt it to our action space using the same \texttt{Qwen2.5-VL-7B} backbone.
    \item \textbf{Embodied-Reasoner \cite{embodied-resoner} (Fine-tuned):} A recent framework extending deep reasoning to embodied settings. We adapt its methodology to our task and data.
    \item \textbf{Heuristic-Search (Rule-Based).} To validate the necessity of learning-based planning, we implement a rigid rule-based agent using the \texttt{Qwen2.5-VL-7B} backbone. This agent follows a fixed \texttt{"Ask-then-Explore"} template: it always asks one clarification question at the start, then navigates to the most likely location inferred from the response. If the target is not found, it explores randomly. This serves as a lower-bound reference for non-adaptive strategies.
\end{itemize}

\paragraph{Ablations}
\begin{itemize}
    \item \textbf{ESearch-R1 w/o Dialogue.} This ablation removes the \texttt{Ask} action to isolate the contribution of active dialogue in resolving ambiguity.
    \item \textbf{ESearch-R1 w/o Memory.} This ablation removes the \texttt{GetMemory} action to quantify the impact of episodic memory on exploration efficiency.
    \item \textbf{ESearch-SFT.} This version uses the policy after Stage I (SFT-CoT) only, without online optimization, to evaluate the necessity of HC-GRPO for refining cost-benefit trade-offs.
\end{itemize}

\paragraph{Evaluation Metrics}
We evaluate performance using the following metrics:
\begin{itemize}
    \item \textbf{Success Rate (SR, \%).} The percentage of episodes where the agent correctly identifies and navigates to the ground-truth target.
    \item \textbf{Total Task Cost (TTC).} The average cumulative cost incurred during successful episodes, directly measuring the agent's efficiency (lower is better):
    \[
        \mathrm{TTC}=\mathbb{E}\!\left[\sum_{t} C(a_t, h_t)\,\middle|\,\text{Success}\right].
    \]
    \item \textbf{Success Weighted by Cost (SwC).} Our primary metric for overall performance, balancing success with cost-efficiency:
    \[
        \mathrm{SwC}=\mathrm{SR}\times\frac{C_{\text{ref}}}{\max(\mathrm{TTC},\,C_{\text{ref}})}.
    \]
    Here, $C_{\text{ref}}=2.0$ is a normalization constant derived from the average optimal trajectory cost on the validation set. This metric rewards agents that achieve high success rates at a low cost.
    \item \textbf{Decision Quality (LLM-based Score).} We employ an LLM-based evaluator (\texttt{Gemini-2.5-Pro}) to score the quality of the agent's chain-of-thought reasoning on a $[0,1]$ scale. The score assesses logical coherence, strategic foresight, and explicit cost-awareness.
\end{itemize}

\subsection{Main Results: Superiority in Collaborative Efficiency}

\begin{table*}[t]
\centering
\caption{Main results under different numbers of distractors ($n$). All values are mean $\pm$ std over five seeds. SR (\%) denotes success rate; TTC denotes average total task cost on successful episodes (lower is better). ``Avg.'' denotes averages over all tasks.}
\label{tab:main}
\begin{tabular}{lccccccc}
\hline
Model & SR $(n{=}1\text{--}2)\,\uparrow$ & TTC $(n{=}1\text{--}2)\,\downarrow$ & SR $(n{=}3\text{--}4)\,\uparrow$ & TTC $(n{=}3\text{--}4)\,\downarrow$ & Avg. SR $\uparrow$ & Avg. TTC $\downarrow$ & SwC $\uparrow$ \\
\hline
Heuristic-Search (7B)         & $19.5\pm2.1\%$ & $4.1\pm0.5$ & $18.2\pm2.5\%$ & $4.3\pm0.6$ & $18.9\pm1.8\%$ & $4.2\pm0.4$  & $0.09\pm0.01$ \\
ReAct (Qwen2.5-VL-7B)         & $21.4\pm2.8\%$ & $3.8\pm0.8$ & $22.4\pm3.2\%$ & $3.7\pm1.0$ & $21.3\pm2.7\%$ & $3.7\pm0.6$ & $0.11\pm0.01$ \\
Ask-To-Act                    & $46.9\pm2.5\%$ & $3.0\pm0.3$ & $41.2\pm2.3\%$ & $3.2\pm0.3$ & $44.0\pm1.8\%$ & $3.1\pm0.2$ & $0.28\pm0.03$ \\
Embodied-Reasoner             & $59.2\pm2.4\%$ & $3.4\pm0.3$ & $55.4\pm2.1\%$ & $3.6\pm0.3$ & $57.3\pm1.5\%$ & $3.5\pm0.2$ & $0.32\pm0.03$ \\
ReAct (Gemini-2.5-Pro)        & $57.7\pm2.1\%$ & $\underline{2.3\pm0.3}$ & $56.2\pm1.8\%$ & \underline{$2.6\pm0.4$} & $56.5\pm1.3\%$ & \underline{$2.5\pm0.2$} & $0.45\pm0.03$ \\
ReAct (Gemini-2.5-Flash)      & $60.6\pm2.1\%$ & $2.5\pm0.3$ & $58.0\pm1.8\%$ & \underline{$2.6\pm0.4$} & $58.9\pm1.3\%$ & $2.6\pm0.2$ & \underline{$0.46\pm0.03$} \\
ReAct (Qwen2.5-VL-32B)        & $\mathbf{63.4\pm4.0\%}$ & $2.9\pm0.2$ & $\underline{58.8\pm1.0\%}$ & $3.6\pm0.2$ & $\underline{60.0\pm2.3\%}$ & $3.3\pm0.1$ & $0.36\pm0.02$ \\
\rowcolor{blue!20}
\textbf{ESearch-R1}           & $\underline{62.9\pm2.0\%}$ & $\mathbf{1.6\pm0.1}$ & $\mathbf{60.0\pm1.9\%}$ & $\mathbf{1.5\pm0.2}$ & $\mathbf{61.5\pm1.6\%}$ & $\mathbf{1.6\pm0.1}$ & $\mathbf{0.59\pm0.02}$ \\

\hline
\end{tabular}
\end{table*}

\paragraph{Robustness Against Ambiguity (RQ1)}
As shown in Table \ref{tab:main}, ESearch-R1 consistently outperforms all baselines in Success Rate (SR). Notably, under high-ambiguity conditions (3–4 distractors), traditional ReAct agents degrade significantly (SR $\approx$ 22\%), struggling to manage uncertainty. In contrast, ESearch-R1 maintains a high success rate (60.0\%), demonstrating its capability to robustly handle under-specified instructions typical in dynamic service environments.

\paragraph{Optimizing the Cost-Efficiency Trade-off (RQ2)}
The most significant advantage of ESearch-R1 lies in its cost efficiency.
\begin{itemize}
    \item Reduced Operational Cost: Compared to the strongest baseline (ReAct Qwen2.5-VL-32B), ESearch-R1 reduces the Total Task Cost (TTC) by approximately 50\% (from 3.3 to 1.6). This reduction directly translates to energy savings and better resource management.
    \item Minimized Human Disturbance: By intelligently utilizing internal GetMemory actions (cost $\approx$ 0) instead of blindly asking questions, ESearch-R1 reduces unnecessary human interactions. This is reflected in the superior SwC score (0.59 vs. 0.36), indicating that our agent achieves success while respecting the user's cognitive budget.
\end{itemize}

\paragraph{Effectiveness of Training Strategy}
Figure \ref{fig:grpo} illustrates the training dynamics of our HC-GRPO algorithm. The steady increase in Mean Critic Reward confirms that the agent effectively learns the heterogeneous cost structure. Crucially, the Mean Response Length decreases over time, suggesting the model learns to generate concise, efficient reasoning chains—a desirable trait for edge-computing deployments where inference latency matters.

\subsection{Cost Parameter Sensitivity Analysis}

\begin{figure*}[t]
    \centering
    \includegraphics[width=\linewidth]{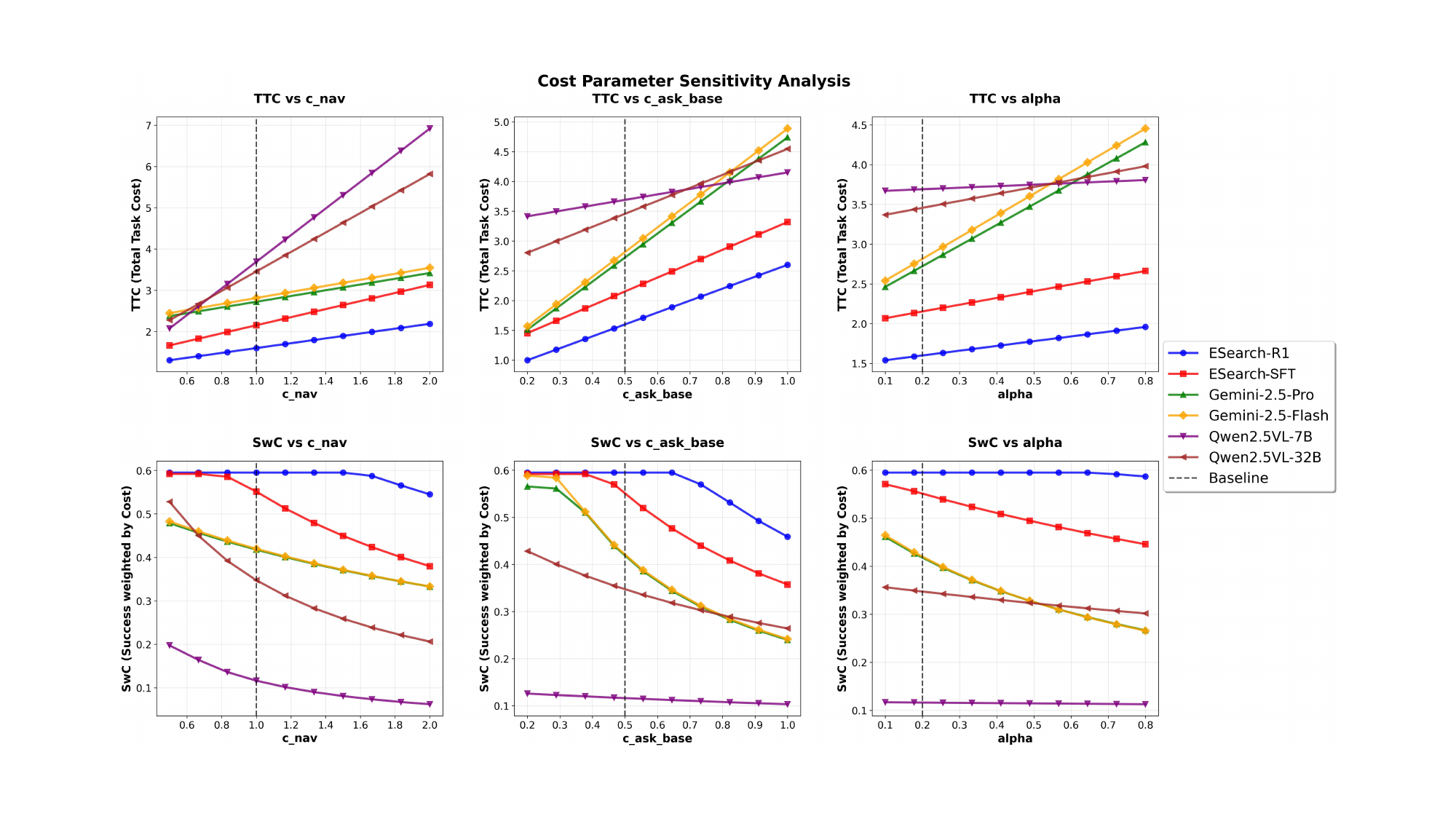}
    \caption{Cost parameter sensitivity analysis. ESearch-R1 (blue) consistently outperforms baselines across all cost configurations, demonstrating robust cost-aware optimization. The vertical dashed lines indicate our default parameter settings.}
    \label{fig:cost_sensitivity}
\end{figure*}

To validate the robustness of ESearch-R1’s cost-aware optimization, we conducted a sensitivity analysis by varying key cost parameters ($c_{\text{nav}}$, $c_{\text{ask\_base}}$, $\alpha$). As shown in Figure~\ref{fig:cost_sensitivity}, ESearch-R1 maintains its superior performance in terms of TTC and SwC across all tested configurations. Even when navigation becomes cheaper or dialogue more expensive, our framework adapts its strategy and consistently identifies more cost-efficient solutions than the baselines. This demonstrates that our cost-aware learning approach is not overfitted to a specific cost structure but learns a generalizable meta-policy for trading off information gain and execution cost, validating our core hypothesis.

\subsection{Dissecting Decision Strategies}
The quantitative efficiency gains are rooted in a fundamental shift in the agent's decision-making strategy, which we analyze below.

\begin{table*}[t]
\centering
\caption{Decision quality and strategic tendencies. ``Logic'' and ``Strategy'' are LLM-based scores in $[0,1]$.}
\label{tab:strategy}
\begin{tabular}{lcccccc}
\hline
Model & Avg. traj. length $\downarrow$ & Logic $\uparrow$ & Strategy $\uparrow$ & Questioning $\uparrow$ & Memory $\uparrow$ & Reasoning $\uparrow$ \\
\hline
Ask-To-Act              & $5.8_{\pm 0.2}$ & $0.60$ & $0.49$ & $\underline{0.87}$ & $0.88$ & $0.48$ \\
ReAct (Qwen2.5-VL-32B)  & $5.6_{\pm 0.0}$ & $\underline{0.75}$ & $0.51$ & $0.86$ & $0.86$ & $\mathbf{0.75}$ \\
ReAct (Gemini-2.5-Flash) & $5.5_{\pm 0.2}$ & $0.56$ & $0.44$ & $0.53$ & $\mathbf{0.93}$ & $0.41$ \\
ReAct (Gemini-2.5-Pro) & $5.5_{\pm 0.2}$ & $0.56$ & $0.43$ & $0.51$ & $ 0.92 $ & $0.42$ \\
ReAct (Qwen2.5-VL-7B)   & $5.2_{\pm 0.6}$ & $0.70$ & $0.20$ & $0.72$ & $0.89$ & $0.51$ \\
Embodied-Reasoner       & $5.1_{\pm 0.2}$ & $0.64$ & \underline{$0.60$} & $0.70$ & $\underline{0.90}$ & $0.63$ \\
\textbf{ESearch-R1}     & $\mathbf{4.3_{\pm 0.1}}$ & $\mathbf{0.76}$ & $\mathbf{0.77}$ & $\mathbf{0.88}$ & $0.80$ & $\underline{0.71}$ \\
\hline
\end{tabular}
\end{table*}

\paragraph{Shorter Decision Chains, Higher Efficiency}
As detailed in Table~\ref{tab:strategy}, ESearch-R1 completes tasks with the shortest average trajectory length (4.3 steps). This reflects its "disambiguate first, then navigate" strategy: by prioritizing low-cost \texttt{Ask} and \texttt{GetMemory} actions to quickly identify the target, it avoids the lengthy trial-and-error physical exploration common in baselines.

\paragraph{Superior Reasoning and Strategic Quality}
The LLM-based evaluation corroborates these behavioral findings. ESearch-R1 achieves the highest scores for logical coherence and strategic quality, indicating that its decisions are not just effective but are also supported by higher-quality, cost-aware reasoning.

\begin{figure}[t]
    \centering
    \includegraphics[width=1\linewidth]{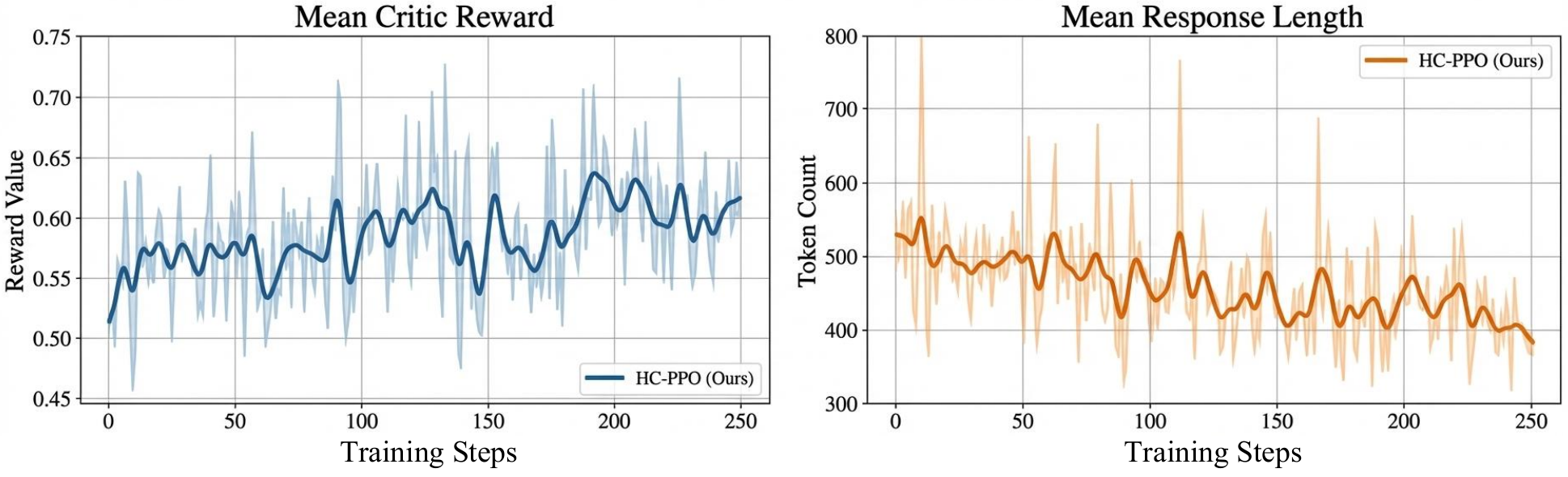}
    \caption{Training dynamics during HC-GRPO’s online optimization. Left: average return (task reward minus total cost) steadily increases. Right: average response token length (including CoT and actions) decreases markedly.}
    \label{fig:grpo}
\end{figure}

\paragraph{Training Validates the Evolution of Strategy}
The training dynamics of HC-GRPO (Figure~\ref{fig:grpo}) reveal how this strategic shift is learned. The average return (reward minus cost) steadily increases, showing the agent is successfully optimizing the cost-aware objective. Concurrently, the average length of the generated CoT decreases, suggesting the agent learns to produce more concise and confident decisions, which also implies lower inference latency.

\begin{figure}[t]
    \centering
    \includegraphics[width=1\linewidth]{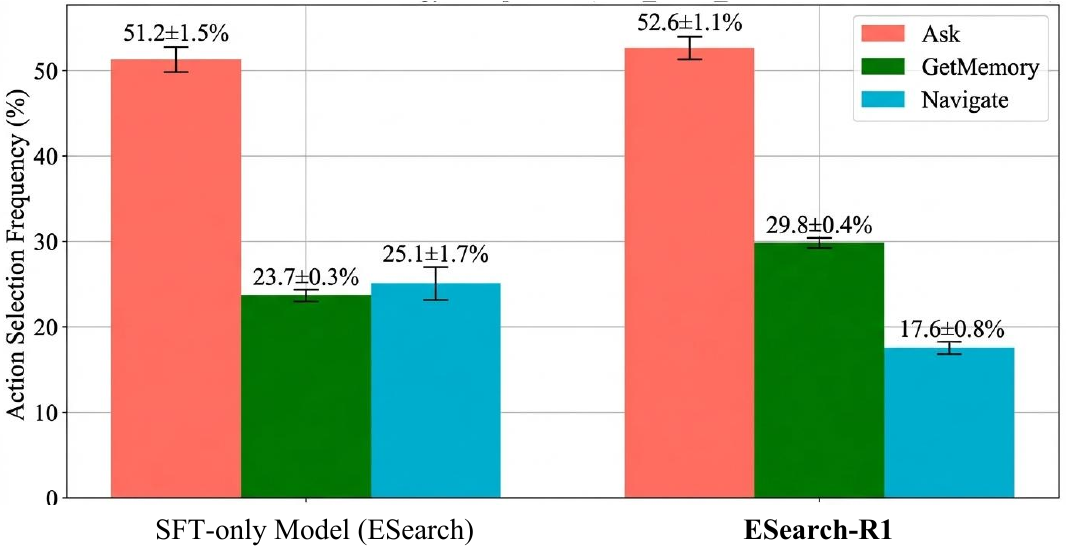}
    \caption{Comparison of decision distributions between ESearch-SFT and ESearch-R1.}
    \label{fig:dist}
\end{figure}

\paragraph{A Strategic Shift in Decision Distribution}
This learned efficiency manifests as a quantifiable shift in the agent's action distribution (Figure~\ref{fig:dist}). Compared to the SFT-only model, the final ESearch-R1 policy increases its use of the zero-cost \texttt{GetMemory} action while reducing its reliance on high-cost \texttt{Navigate} actions. This strategic substitution—replacing expensive physical exploration with low-cost internal information retrieval—is the primary driver of its superior cost-efficiency.

\noindent\textbf{Synthesis.} The online optimization stage of HC-GRPO refined a model that “knows what to do” (from SFT) into one that “knows how to do it at minimum cost.” The cost-aware reward function encouraged the agent to proactively substitute some high-cost \texttt{Navigate} actions with zero-cost \texttt{GetMemory} when beneficial. This strategic shift—from expensive physical exploration to low-cost internal information retrieval—explains why ESearch-R1 simultaneously increased SR (from 59.2\% to 61.5\%) and reduced TTC by nearly 25\% (from 2.4 to 1.8), validating our two-stage training paradigm of “inject logic first, then optimize trade-offs.”

\subsection{Ablation Studies: Validating Core Components}
To validate the contributions of each core component of our framework, we conducted a series of ablation studies. The results, summarized in Table~\ref{tab:ablation}, underscore the indispensable roles of dialogue, memory, and our online optimization strategy.

\begin{table}[t]
\centering
\caption{Ablation results for ESearch-R1.}
\label{tab:ablation}
\begin{tabular}{lccc}
\hline
Ablation & SR$\uparrow$ & TTC$\downarrow$ & SwC$\uparrow$ \\
\hline
ESearch-R1 (Full) & $\mathbf{61.5\%}$ & $\underline{1.6}$ & $\mathbf{0.59}$ \\
w/o Dialogue (Ask) & $10.5\%$ & $\mathbf{1.0}$ & $0.1$ \\
w/o Memory (GetMemory) & $52.0\%$ & $2.3$ & $0.52$ \\
w/o HC-GRPO & $\underline{59.2\%}$ & $2.3$ & $\underline{0.53}$ \\
\hline
\end{tabular}
\end{table}

\paragraph{Necessity of Dialogue and Memory}
\begin{itemize}
    \item \textit{Removing dialogue (w/o Dialogue).} Disabling the \texttt{Ask} action causes the SR to plummet to just 10.5\%, as the agent loses its primary tool for resolving target ambiguity. This result confirms that \textbf{active dialogue is indispensable for solving tasks with under-specified instructions.}
    \item \textit{Removing memory (w/o Memory).} Without access to episodic memory, the agent's SR drops to 52.0\% and its TTC increases significantly. This demonstrates that the agent is forced into redundant physical exploration to re-verify information it has already seen. Thus, \textbf{episodic memory is a critical component for cost-efficient exploration.}
\end{itemize}

\paragraph{Value of Online Optimization with HC-GRPO}
Comparing the full ESearch-R1 model with its SFT-only counterpart reveals the crucial role of online optimization. While ESearch-SFT learns foundational decision logic from expert data (SR 59.2\%), it remains suboptimal in its cost trade-offs (TTC 2.3). The HC-GRPO optimization stage is essential for refining this policy, simultaneously improving SR to 61.5\% and reducing TTC to 1.6. This confirms that direct, cost-sensitive interaction with the environment is necessary to master the complex trade-offs required for true efficiency.

\subsection{Qualitative Case Studies}
\paragraph{Success Case}
In a scene with multiple screwdrivers, ESearch-R1 first issued an \texttt{Ask} action (cost 0.5) to learn the target was "Phillips." It then used a zero-cost \texttt{GetMemory} action to recall seeing a Phillips screwdriver on the workbench. Finally, it executed a single, short \texttt{Navigate} action, successfully completing the task with minimal total cost. This trajectory exemplifies how a sequence of low-cost information actions can effectively replace expensive physical search.

\paragraph{Failure Case}
Faced with multiple bottles, the agent asked one question but then navigated prematurely without seeking further clarification or checking its memory. This led to an incorrect final choice. This highlights an area for future work in improving the agent's confidence estimation and risk assessment when dealing with ambiguous user responses.

\section{Conclusion}

In this work, we presented ESearch-R1, a cost-aware embodied decision-making framework designed to resolve target ambiguity in complex embodied settings. By unifying dialogue, memory retrieval, and physical exploration into a single POMDP formulation and optimizing it via our HC-GRPO algorithm, the agent learns to balance information gain against heterogeneous execution costs. Extensive experiments on ESearch-Bench demonstrate that ESearch-R1 achieves state-of-the-art performance, significantly reducing navigation costs while maintaining high success rates compared to existing baselines.

Despite these promising results, several limitations remain. First, while our reasoning module is robust, perception failures (e.g., detection errors under occlusion) can constrain overall success rates. Second, our current cost function uses static coefficients, whereas real-world scenarios often involve dynamic priorities (e.g., urgency vs. energy saving). Third, the inference latency of the 7B-parameter MLLM may impact real-time responsiveness in highly dynamic collaborative tasks. In future work, we will incorporate uncertainty-aware perception and explore dynamic cost learning to adapt to varying task urgencies. Furthermore, we intend to optimize the MLLM backbone for edge computing to enhance real-time performance.

\appendices

\bibliographystyle{IEEEtran}
\bibliography{references}

\end{document}